%% file: main_camera_ready.tex
\documentclass[10pt,twocolumn,letterpaper]{article}

\usepackage[pagenumbers]{cvpr} % To force page numbers, e.g. for an arXiv version

\usepackage{multirow}
\usepackage{gensymb}

\usepackage[table]{xcolor} % Load xcolor with table option
\definecolor{LightGray}{gray}{0.9} % Define a custom light gray color

\newcolumntype{G}{>{\columncolor{LightGray}}c} % Define a new column type 'G' for a centered column with LightGray background

\usepackage{listings}
\definecolor{cvprblue}{rgb}{0.21,0.49,0.74}
\usepackage[pagebackref,breaklinks,colorlinks,allcolors=cvprblue]{hyperref}

\def\paperID{****} % *** Enter the Paper ID here
\def\confName{CVPR}
\def\confYear{2026}

\title{Direction-aware 3D Large Multimodal Models}

\author{Quan Liu$^1$ \ \ \ \ \ Weihao Xuan$^{2,3}$ \ \ \ \ \ Junjue Wang$^2$ \ \ \ \ \ Naoto Yokoya$^{2,3}$ \ \ \ \ \ Ling Shao$^4$ \ \ \ \ \ Shijian Lu$^{1,\dagger}$ \\
$^1$Nanyang Technological University \ \ \ \ \ $^2$The University of Tokyo \ \ \ \ \ $^3$RIKEN AIP\\
$^4$UCAS-Terminus AI Lab, University of Chinese Academy of Sciences
}

\begin{document}
\maketitle
\input{sec/0_abstract_v4}
\let\thefootnote\relax
\footnotetext{\textsuperscript{\textdagger} Corresponding author: \href{shijian.lu@ntu.edu.sg}{Shijian.Lu@ntu.edu.sg}}
\footnotetext{~~~Code is available at \href{https://github.com/liuQuan98/PoseAlign3D}{https://github.com/liuQuan98/PoseAlign3D}}
\input{sec/1_intro_v6}

\input{sec/2_related_work_v1}
\input{sec/3_data_pipeline_v2}
\input{sec/4_method_v1}
\input{sec/5_results_v2}

\input{sec/6_conclusion}

\input{sec/acknowledgment}
{
    \small
    \bibliographystyle{ieeenat_fullname}
    \bibliography{main}
}

\input{sec/7_appendix_v2}

\end{document}

%% file: sec/0_abstract_v4.tex
\begin{abstract}
3D large multimodal models (3D LMMs) rely heavily on ego poses for enabling directional question-answering and spatial reasoning. However, most existing point cloud benchmarks contain rich directional queries but lack the corresponding ego poses, making them inherently ill-posed in 3D large multimodal modelling. In this work, we redefine a new and rigorous paradigm that enables direction-aware 3D LMMs by identifying and supplementing ego poses into point cloud benchmarks and transforming the corresponding point cloud data according to the identified ego poses. We enable direction-aware 3D LMMs with two novel designs. The first is \textit{PoseRecover}, a fully automatic pose recovery pipeline that matches questions with ego poses from RGB-D video extrinsics via object–frustum intersection and visibility check with Z-buffers. The second is \textit{PoseAlign} that transforms the point cloud data to be aligned with the identified ego poses instead of either injecting ego poses into textual prompts or introducing pose-encoded features in the projection layers. 
Extensive experiments show that our designs yield consistent improvements across multiple 3D LMM backbones such as LL3DA, LL3DA-SONATA, Chat-Scene, and 3D-LLAVA, improving ScanRefer mIoU by $30.0\%$ and Scan2Cap LLM-as-judge accuracy by $11.7\%$. In addition, our approach is simple, generic, and training-efficient, requiring only instruction tuning while establishing a strong baseline for direction-aware 3D-LMMs.
\end{abstract}

%% file: sec/1_intro_v6.tex
\section{Introduction}
\label{sec:intro}
Generalist point cloud 3D large multimodal models (3D-LMMs) pursue broad competence in 3D scene understanding and reasoning, targeting key tasks such as object grounding \cite{yang2021sat, yuan2022toward,huang2022multi,wang2024embodiedscan}, referring \cite{chen2020scanrefer, achlioptas2020referit3d,yuan2021instancerefer}, question answering \cite{azuma2022scanqa, yan2023comprehensive}, ego position reasoning \cite{ma2022sqa3d}, object captioning \cite{chen2021scan2cap,yuan2022x}, route planning \cite{savva2019habitat}, segmentation \cite{schult2022mask3d,he2024segpoint}, detection \cite{shen2023v}, etc. Such enabling represents a key step towards the visual-cognitive core of future embodied agents \cite{huang2023embodied, deng20253d}. In particular, an ideal generalist 3D-LMM should possess persistent awareness of ego poses and spatial directions in the world coordinate for accurate understanding of 3D input data \cite{qian2024x,wu20243d}, robust and alias-free spatial reasoning \cite{yuan2025empowering,deng20253d}, and safely navigating in 3D environments \cite{huang2023embodied}.

\begin{figure}[t]
  \centering
   \includegraphics[width=0.99\linewidth]{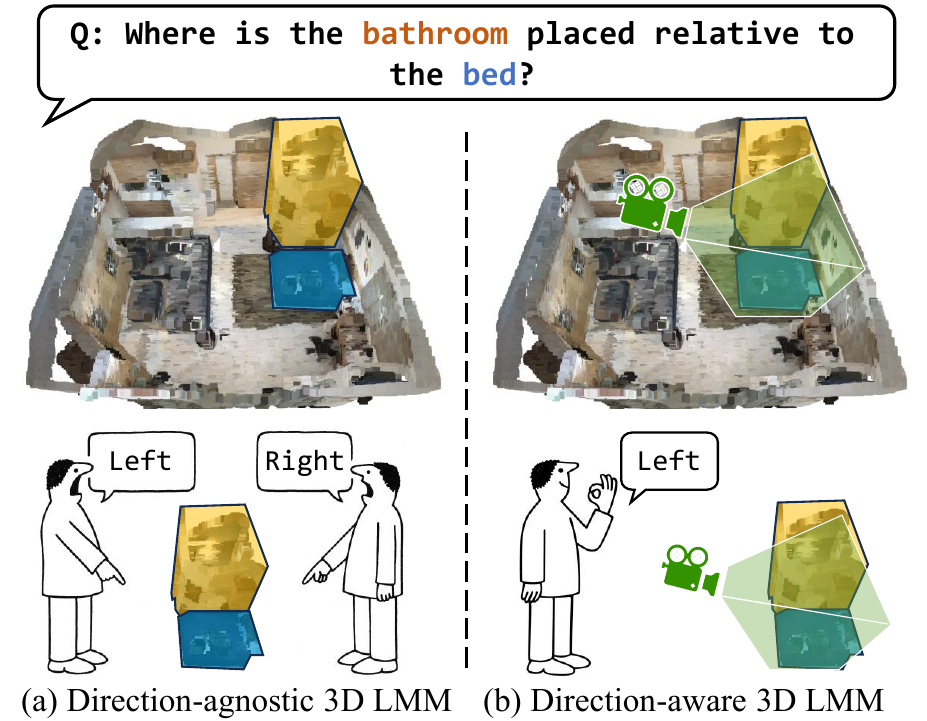}
    % \vspace{-0.3cm}
   \caption{Ego pose is critical in spatial reasoning and understanding. (a) Direction-agnostic 3D LMMs are struggling to reason spatial directions due to the absence of ego-pose information. (b) Incorporating ego pose resolves directional ambiguity, enabling consistent and robust spatial reasoning.}
   \label{fig:intro}
   \vspace{-0.3cm}
\end{figure}

Ego-pose awareness entails distinguishing lateral directions (e.g., `left', `right', `front', `back') that are defined according to a reference frame \cite{freksa1998spatial}. Two forms of directions have been widely adopted: 1) \textit{Egocentric Direction} that defines spatial relations relative to the agent's own body axes (\textit{e.g.}, `the object is on my left') and 2) \textit{Allocentric Direction} that defines spatial relations relative to an external anchor (\textit{e.g.}, 
% `The man is on the left of his car'). 
`the man is on the left of his car'). 
For indoor scenarios where the anchor has no canonical axes, like a plate or a table, the reference of the allocentric direction falls back to the ego agent \cite{freksa1998spatial}. Hence, an ego pose is essential to reconcile the two forms of directions to achieve consistent spatial reasoning and understanding by 3D LMMs. 

However, most existing 3D indoor datasets such as ScanRefer \cite{chen2020scanrefer}, Multi3DRefer \cite{zhang2023multi3drefer}, ScanQA \cite{azuma2022scanqa}, Scan2Cap \cite{chen2021scan2cap}, and Nr3D \cite{achlioptas2020referit3d} contain many allocentric directional queries but provide no explicit ego poses, positioning spatial reasoning ill-posed and hindering the acquisition of persistent ego pose awareness as illustrated in Figure \ref{fig:intro}. 
This omission largely stems from their assumption of a global third-person rather than an egocentric first-person perspective of the scene.
Prior studies address this omission mostly by creating new datasets instead of rectifying existing ones, requiring 3D LMMs to infer ego poses while responding to spatial queries \cite{ma2022sqa3d,yuan2025empowering}. Actually, inferring ego poses as a latent variable is conceptually redundant, as ego poses are already available when an embodied agent collects indoor point clouds via simultaneous localization and mapping (SLAM) \cite{murai2025mast3r}. 
To this end, we propose to directly incorporate the readily available pose data as model inputs due to two factors:
(1) it rigorously resolves the directional ambiguity in existing 3D indoor benchmarks; (2) it does not affect the application scope of 3D LMMs by leveraging the `free-lunch' pose in the practical embodied workflow. More discussion is available in section \ref{sec:annotation_acquisition}.

For efficient implementation, we enable the new paradigm with two new designs. The first is \textit{PoseRecover}, an automatic pose generation pipeline that addresses the lack of ego poses in existing benchmarks. Specifically, \textit{PoseRecover} retrieves question-related camera poses from ScanNet RGB-D sequences by matching camera frustums with relevant object annotations such as segmentation masks, bounding boxes, and location annotators, leading to a list of candidate camera poses for online training or inference. During the forward pass, candidates undergo various screenings to purge opposite-view errors and ensure authenticity while maintaining a certain degree of variety. The second is \textit{PoseAlign}, a simple solution that incorporates the identified ego pose into point cloud data to enable their interpretation by existing 3D LMMs. Specifically, \textit{PoseAlign} repositions point clouds to align with the identified ego poses, enabling universal direction-awareness across existing 3D LMMs of different architectures. Thanks to the coordinate sensitivity of pretrained point cloud encoders, the direction awareness can be boosted without even tuning the encoder of 3D LMMs.

The contributions of this work can be summarized in three major aspects. 
First, we identify that most existing 3D LMM benchmarks suffer from ill-posed directions and propose a new paradigm that mitigates this problem effectively by incorporating an ego pose.
Second, we design PoseRecover and PoseAlign, the former being a pose generation pipeline that addresses the ill-posed problem in 3D LMM benchmarks by recovering mission-critical pose data, and the latter being a simple yet effective modifier that enables direction-awareness for existing 3D LMMs by injecting pose data into the point cloud.
Third, extensive experiments show that our approach improves the direction awareness substantially and consistently across 4 different benchmarks and all 3D LMM architectures.

%% file: sec/2_related_work_v1.tex
\section{Related Work}
\label{sec:related_work}

\subsection{Indoor 3D LMM Benchmarks}

Following the huge success of 2D visual understanding benchmarks \cite{liu2023visual,liu2024improved} and the richly-annotated indoor datasets \cite{dai2017scannet,chang2017matterport3d,wang2024embodiedscan}, the first batch of 3D understanding benchmarks \cite{achlioptas2020referit3d,yan2023comprehensive,yuan2021instancerefer} take similar approaches to link objects with text descriptions in 3D with expert human annotators, achieving a stunning variety and volume of tasks, \textit{e.g.}, referral \cite{chen2020scanrefer}, captioning \cite{chen2021scan2cap}, and question-answering \cite{azuma2022scanqa}. While one could argue that their human-centric pipelines implied the viewing direction must be from a human standing inside the room, not explicitly defining such poses causes their questions to be inherently ill-posed \cite{yuan2025empowering}. Often completed by crowdsourcing, these benchmarks could never again accurately recover those poses.

Later attempts to improve directional awareness \cite{huang2022multi,ma2022sqa3d,guo2023viewrefer,fu2024scene,wang2024embodiedscan,yuan2025empowering} have recognized this problem but failed to fully address it. SQA3D \cite{ma2022sqa3d} provides a text description of an ego situation and a QA pair that requires spatial reasoning. 
While the text description implicitly conveys the ego pose, this approach introduces textual ambiguity and is not practically meaningful in real-world applications, where ego pose descriptions are not always available.
View2Cap \cite{yuan2025empowering} proposes the Situation Grounding module, which takes in a frustum-cropped partial point cloud and regresses the associated camera pose. However, pose regression is far from textual reasoning with direction awareness, thus its effect is limited. 
Scene-LLM \cite{fu2024scene} enhances egocentric awareness using a two-step inference scheme, with first an egocentric frustum-cropped point cloud and then a scene-level point cloud. However, this approach is unnecessarily complex and uses proprietary code and data.
Furthermore, these methods all set new benchmarks rather than altering existing ones, which still suffer from the problem of ill-posed directional definition.

\subsection{3D LMMs}

\begin{figure*}[t]
  \centering
   \includegraphics[width=0.95\linewidth]{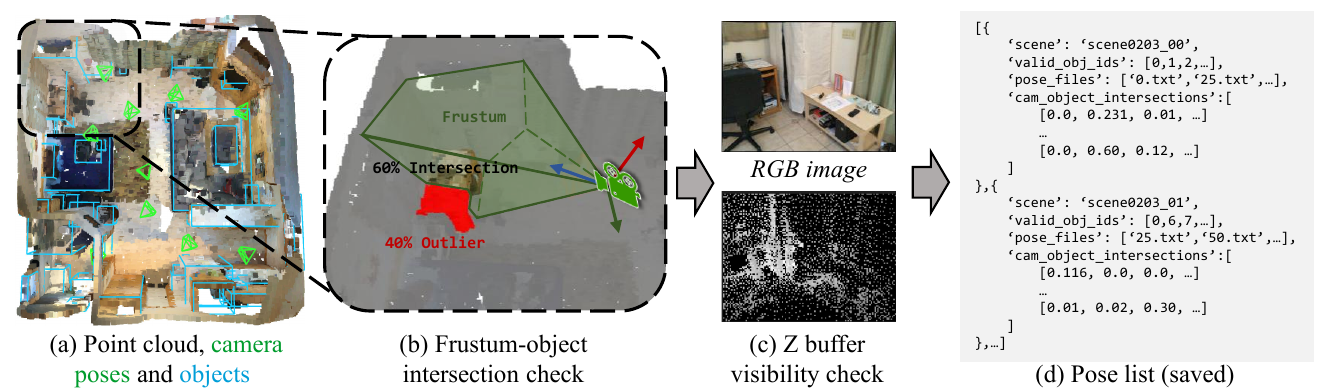}
   \vspace{-0.2cm}
   \caption{\textbf{The offline data generation pipeline for PoseRecover}. (a) Object annotations and camera poses are obtained from ScanNet-v2 \cite{dai2017scannet}. Camera poses and objects are downsampled for visibility. Zoom in for details. (b) PoseRecover exhaustively calculates the intersection rates between objects and camera frustums. (c) Visibility of the intersection is further validated with a z-buffer. (d) These intersection rates are saved and later sampled during training or inference to supplement ego poses to models.}
   \label{fig:data_pipeline}
\end{figure*}

\paragraph{2D-based 3D LMMs.} Based on powerful 2D-pretrained networks, \textit{e.g.}, CLIP \cite{radford2021learning}, DINOv2 \cite{oquab2023dinov2}, and LLAVA-Video \cite{zhang2024video}, this line of research focuses on projecting RGB-D image features back to 3D for understanding \cite{hong20233d,fu2024scene,zheng2025video,wu2025spatial}. 3D-LLM \cite{hong20233d} explored a mixture of projection methods, including direct projection, SLAM \cite{murthy2019gradslam}, and NeRF \cite{mildenhall2021nerf}. 
Video-3D-LLM \cite{zheng2025video} and Spatial-MLLM \cite{wu2025spatial} parse the scene as an RGB-D video with pretrained video- or vision-language-models (VLMs). These methods introduce implicit information of ego pose in the images, therefore can be seen as natural implementations of our method.

\paragraph{Detector-based 3D LMMs.} To reconcile between computational complexity and loss of task-centric information, these methods choose 3D detection \cite{shen2023v} or instance segmentation \cite{schult2022mask3d} to encode objects as LLM input tokens \cite{huang2024chat,wang2023chat,huang2023embodied,chen2024grounded,he2024segpoint,zhu20233d}. 
For example, Chat-Scene \cite{huang2024chat} extracts object-centric features via a pretrained detector and passes these object features as tokens using a unified object identifier token. However, these methods are inherently limited by the detector capability and ignore background information, which is essential for reasoning in real scenarios.

\paragraph{3D-backbone-based 3D LMMs.} This line of method processes the scene-level point cloud in a single forward pass which is then compressed and sent to LLM for reasoning \cite{chen2024ll3da,deng20253d,huang2025reason3d}. 
LL3DA \cite{chen2024ll3da} encodes scene-level features through a Q-Former \cite{li2023blip} for natural human interaction.
Inspired by superpoint transformers \cite{kolodiazhnyi2024oneformer3d,lai2023mask,sun2023superpoint}, 3D-LLAVA \cite{deng20253d} completes point cloud processing and compression simultaneously using the cluster centers, \textit{i.e.}, superpoints, as tokens.
Due to natural processing of scene contexts, these methods achieve state-of-the-art performance but are highly contingent on the point cloud encoder design.

%% file: sec/3_data_pipeline_v2.tex
\section{Data Pipeline}
\label{sec:data_pipeline}

Existing point cloud benchmarks, including ScanQA, ScanRefer, Multi3DRefer, SQA3D, Scan2Cap, and Nr3D, contain 40\% to 95\% direction-critical queries according to our analysis, yet they omit the camera pose information required to determine the ego agent’s directional context.
As analyzed in Section 1, this omission renders a significant portion of questions ill-posed with respect to egocentric or allocentric direction understanding.
To address this, we introduce PoseRecover, a fully automatic pipeline that recovers ego poses for all text–scene pairs.
We first introduce the question analysis to establish our motivation, then introduce the offline PoseRecover pipeline as depicted in Figure \ref{fig:data_pipeline}, which matches the question-specific ground-truth object annotation with the camera frustums, and finally introduce two pose selection strategies during online training or inference.

\subsection{Question-Level Analysis}
\label{sec:question_levek_analysis}
We first quantify the degree of directional ambiguity in existing datasets.
An open-source large language model (GPT-OSS-20B \cite{openai2025gptoss120bgptoss20bmodel}) is prompted with a series of question–answer pairs and tasked to determine whether answering the question requires explicit lateral direction reasoning (e.g., “left”, “behind”, “in front of”) \cite{wei2024measuring}. We kindly refer readers to \ref{sec:appendix_direction_gpt_prompt} for the full prompt and \ref{sec:appendix_direction_critical_judgement} for detailed results.
This binary judgment establishes the \textit{direction-critical subsets} of existing datasets, which guide the downstream pose recovery stage.
Across datasets, we find that between 40\% to 95\% of the questions require directional reasoning and are therefore ill-posed without ego-pose supervision.

\subsection{Pose Supplementation via PoseRecover}

Given the intrinsic and extrinsic parameters of the raw RGB-D sequences in ScanNet-v2, we compute the 3D camera frustums (a visible pyramid with near and far cutoff) and measure their spatial intersection with question-related objects, and save all normalized intersection values into a pose list.
Multiple forms of object annotations are supported:

\begin{figure*}[t]
  \centering
   \includegraphics[width=0.9\linewidth]{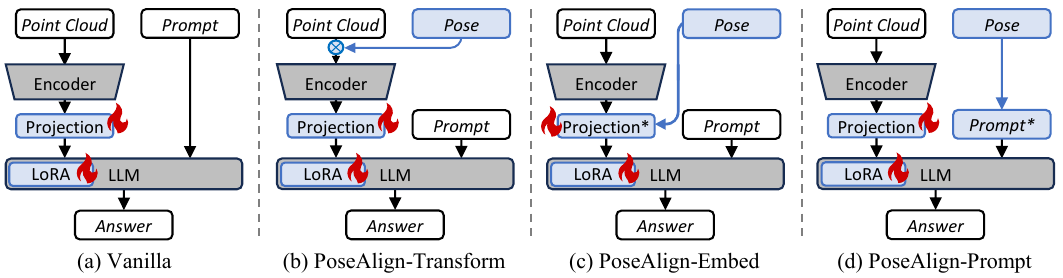}
   \vspace{-0.2cm}
   \caption{\textbf{Three viable designs for PoseAlign}. We explore three mutually exclusive designs to incorporate ego poses into the vanilla model in (a): 1) PoseAlign-Transform that shifts point clouds to the ego reference frame in (b); 2) PoseAlign-Embed that encodes ego poses into point cloud features in (c); 3) PoseAlign-Prompt that integrates ego poses into the text prompt in (d). The projection layer and the LoRA \cite{hu2022lora} weights of the LLM are trained with instruction-tuning.}
   \label{fig:pipeline}
   \vspace{-0.2cm}
\end{figure*}

\paragraph{Segmentation-based.} For segmentation masks, every point in the point cloud is projected into the image plane using the camera intrinsic and extrinsic, forming a Z-buffer following Equations \ref{eq:cam_transform}-\ref{eq:depth_buffer}:
\begin{align}
    &\begin{aligned}
    \qquad (x'_i, y'_i, z'_i)^\intercal=\mathcal{R}^{-1}(p_i-t),
    \label{eq:cam_transform}
    \end{aligned}\\
    &\begin{aligned}
    \qquad (u_i,v_i,1)^\intercal=\lfloor K(x'_i,y'_i,z'_i)^\intercal/z'_i \rfloor,
    \label{eq:intrinsic_transform}
    \end{aligned}
\end{align}

\begin{align}
    &\begin{aligned}
    \qquad  &Z_{\mathcal{P}}^{u_i,v_i} =\min_{j|(u_j,v_j)=(u_i, v_i)}{(z'_j)},\\
    \qquad &\text{s.t.}\ p_i,p_j \in \mathcal{P}, 0\leq u_i < U , 0\leq v_i< V,
    \label{eq:depth_buffer}
    \end{aligned}
\end{align}
where $Z_{\mathcal{P}}\in \mathbb{R}^{U\times V}$ is the Z-buffer, $\mathcal{P}\in \mathbb{R}^{n\times 3}$ is the point cloud, $K, (\mathcal{R}|t)$ are the intrinsic and extrinsic, $U, V$ are the Z-buffer sizes, and $\lfloor\cdot\rfloor$ is the floor operation.
A visibility mask is established by depth-buffer comparison, where only the visible points are counted:
\begin{equation}
    \phi_{seg} = \frac{1}{|M_{obj}|}\sum_{k\in M_{obj}, 0\leq u_k<U, 0\leq v_k<V}\mathbb{I}\big[z'_k < Z_{\mathcal{P}}^{u_k,v_k}+\delta\big]
    \label{eq:seg_intersection}
\end{equation}
where $M_{obj}$ is the segmentation mask consisting of point indices for the object, the margin $\delta=10^{-6}$ tolerates numerical instability, and $\mathbb{I}[\cdot]$ is the Iverson Bracket. The intersection ratio $\phi_{seg}$ is then defined as the proportion of visible object points lying within the frustum. 

\paragraph{Bounding-box-based.} As closed-form intersection between bounding box and frustum can be hard to calculate, we apply Monte-Carlo sampling in the box for an estimation of the intersection ratio $\phi_{box}$ with Equation \ref{eq:box_intersection}:
\begin{align}
    &\begin{aligned}
    &\phi_{box} = \frac{1}{|\mathcal{P}_{sample}|} \sum_{p\in \mathcal{P}_{sample}}{\mathbb{I}[p\in F]},\\
    &\text{s.t.}\ \mathcal{P}_{sample} = \mathrm{meshgrid}(h, w, l, \Delta) + \epsilon ,
    \label{eq:box_intersection}
    \end{aligned}
\end{align}
where $F$ is the frustum region, $h,w,l$ are the bounding box sizes, $\Delta$ is a predefined grid size, $\mathrm{meshgrid}(\cdot)$ is a function that generates uniform spatial grid, and $\epsilon\in [0,\Delta)^{\frac{h}{\Delta}\times \frac{w}{\Delta}\times \frac{l}{\Delta}}$ are random noises.

\begin{figure}[t]
  \centering
   \includegraphics[width=\linewidth]{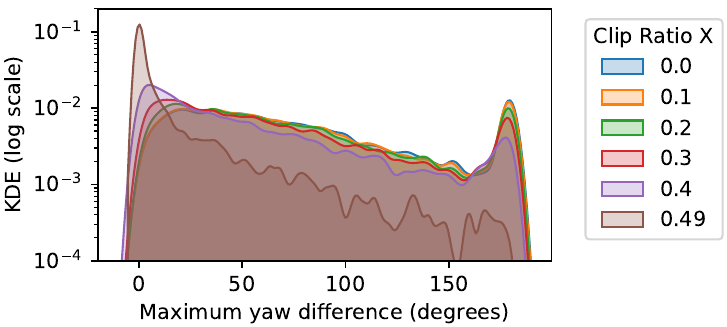}
   \vspace{-0.7cm}
   \caption{\textbf{Effect of the pose clipping.} The KDE \cite{turlach1993bandwidth} of maximum yaw difference among pose candidates rapidly concentrates around zero with increasing clip ratio in ScanQA. Higher clip ratio reduces data variety but boosts pose stability.}
   \label{fig:clip_yaw}
   \vspace{-0.6cm}
\end{figure}

\paragraph{Point-based.} For datasets that provide only a single object location \cite{chen2024grounded}, we compute the normalized pixel distance $\phi_{point}$ between the projected object pixel and the image center following Equations \ref{eq:point_coords}-\ref{eq:point_intersection}, where smaller distances imply larger intersections:
\begin{align}
    &\begin{aligned}
         (u_{obj},v_{obj},1)^\intercal=\lfloor K\mathcal{R}^{-1}(p_{obj}-t)/z'_{obj} \rfloor
         \label{eq:point_coords}
    \end{aligned}\\
    &\begin{aligned}
        \phi_{point} = \begin{cases}\small{1-}\frac{\big\|(\frac{U}{2} - u_{obj}, \frac{V}{2} - v_{obj})\big\|_2}{\big\|\big(\frac{U}{2} , \frac{V}{2}\big)\big\|_2} & \substack{\text{if } 0\leq u_{obj}<U,\\
        \text{and}\ 0\leq v_{obj} < V} \\
        0 & \text{else}
        \end{cases}
        \label{eq:point_intersection}
    \end{aligned}
\end{align}
where $p_{obj}\in \mathbb{R}^3$ is the object center coordinate, and $z'_{obj}$ is the camera-view depth following Equation \ref{eq:cam_transform}.

% Please refer to \ref{sec:appendix_intersection_def} for formal definitions of these procedures. 
These intersection ratios are exhaustively calculated between all camera poses and objects, yielding a camera-object intersection matrix for each scene as depicted in Figure \ref{fig:data_pipeline}(d). Thanks to vectorized implementation, the visibility check costs less than 40 minutes for ScanNet-v2. We refer readers to section \ref{sec:annotation_acquisition} for more discussions.

\subsection{Pose Selection Strategy}

During the forward pass, we retrieve the question-specific column of the camera-object intersection matrix using the ground-truth object annotation in existing datasets. Among the candidate camera poses with non-zero intersection, we evaluate two strategies:
(1) \textbf{Top}: Selecting the pose with the highest intersection ratio, and
(2) \textbf{Clip}: Random sampling after discarding the top and bottom $X\%$ non-zero scores, where $X$ is the clip ratio.
Empirically, the second method with a mediocre $X$ performs better, as it reduces outlier cases like opposite views with $180$\textdegree\ yaw difference among candidate poses while still retaining decent variety, as shown in Figure \ref{fig:clip_yaw}.
This final camera pose serves as the ego reference frame for the language query, enabling an authentic definition of direction semantics across all datasets.

%% file: sec/4_method_v1.tex
\section{Method}
\label{sec:method}

% \subsection{Overview}

% We next examine how the recovered ego pose can be integrated into point cloud LMM frameworks.
Owing to the absence of ego-pose information in existing benchmarks, current 3D LMMs are not designed to accept pose data as direct input. To bridge this gap, we design and compare three PoseAlign variants as displayed in Figure \ref{fig:pipeline}, each augmenting a common backbone (e.g., 3D-LLAVA-style architecture) with ego-pose information through a different pathway:
(1) point cloud transformation (PoseAlign-Transform),
(2) projection-layer positional embedding (PoseAlign-Embed), and
(3) pose prompt injection (PoseAlign-Prompt).
All variants finetune the multimodal projector and the LLM (if applicable) jointly under a LoRA-based adaptation scheme.

\subsection{PoseAlign-Transform}

This variant directly transforms the input point cloud into the recovered camera coordinate frame:
\begin{align}
    &\begin{aligned}
    &\qquad (\mathcal{P}_{aligned}|1)^\intercal = \mathcal{U}\mathcal{T}^{-1}(\mathcal{P}|1)^\intercal,\\
    &\text{s.t.}\ \mathcal{T} = \begin{pmatrix}
                \mathcal{R} & t \\
                0 & 1
                \end{pmatrix},
          \mathcal{U} = \begin{pmatrix}
                0 & 0 & 1 & 0\\
                -1 & 0 & 0 & 0\\
                0 & -1 & 0 & 0\\
                0 & 0 & 0 & 1
                \end{pmatrix},
    \label{eq:posealign_transform}
    \end{aligned}
\end{align}
where $\mathcal{P}$ is the point cloud, $(\mathcal{R}|t)$ is the camera extrinsic matrix, and $\mathcal{U}$ transforms right-down-front camera coordinate to front-left-up coordinate for alignment with the pretrained point cloud encoder.
This operation aligns the spatial distribution of scene points with the ego-viewing direction, ensuring that “left” and “right” are consistently expressed in the egocentric frame.
Because the transformation preserves all geometric relationships, no architectural modifications are required downstream.
This approach achieves the best empirical performance and is therefore used as our default method.

\subsection{PoseAlign-Embed}

Here, we keep the point cloud unchanged but modulate the projection layer of the multimodal projector with an embedding of the 6-DoF pose:
\begin{align}
    &\begin{aligned}
    & f_{aligned} = f + \mathrm{MLP}(\mathrm{encode}(\mathcal{R}, t, \mathcal{P}_f)),
    \label{eq:posealign_embed}
    \end{aligned}
\end{align}
where we encode a series of manual geometric features of the coordinates of the tokens $\mathcal{P}_{f}\in \mathbb{R}^{\mathrm{\#tokens\times 3}}$ into the original feature vectors $f\in \mathbb{R}^{\#tokens\times d}$ using an MLP.
This introduces pose awareness at the feature-projection stage without altering geometry.
% However, it increases sensitivity to representation noise and degrades performance slightly compared with geometric transformation.

\subsection{PoseAlign-Prompt}

In this variant, the pose is serialized as numeric tokens and prepended to the textual prompt in the following format: \texttt{f`position\{t\}, up\{R[:,2]\}, front\{R[:,0]\}, left\{R[:,1]\}'}. This allows the LLM to reason explicitly about spatial direction in text.
Despite its conceptual simplicity, the numeric-token format introduces numeric tokenization overhead and inconsistent positional grounding, resulting in modest performance deterioration. More discussion is available in Section \ref{sec:appendix_transform_superiority_discussion}

%% file: sec/5_results_v2.tex
\section{Results}
\label{sec:results}

\subsection{Experiment Setup}

\paragraph{Datasets.}
We conduct experiments on a series of datasets including ScanRefer \cite{chen2020scanrefer}, Multi3DRefer \cite{zhang2023multi3drefer}, ScanQA \cite{azuma2022scanqa}, SQA3D \cite{ma2022sqa3d}, and Scan2Cap \cite{chen2021scan2cap}. These datasets are based on the ScanNet-v2 dataset \cite{dai2017scannet}, consisting of 1,513 indoor scenes, among which 1,201 are used for training and 312 are used for validation. 

\paragraph{Metrics.}
We adopt both traditional metrics and new metrics of LLM-as-judge \cite{wei2024measuring} for a comprehensive assessment suite. The LLM-as-judge Accuracy (\textbf{L-A}) adopts GPT-OSS-20B \cite{openai2025gptoss120bgptoss20bmodel} or GPT-5-mini \cite{openai2025chatgpt5} (see section \ref{sec:gpt_5_llm_as_judge}) as the base model. The prompt (see section \ref{sec:prompts}) is based on the SimpleQA benchmark \cite{wei2024measuring}. L-A is consistent under different paraphrases of the same meaning, focusing on the invariant meanings instead of more random wording choices \cite{wei2024measuring}. We also report traditional metrics including CiDEr (\textbf{C}), BLEU-4 (\textbf{B-4}), METEOR (\textbf{M}), and ROUGE-L (\textbf{R}) for QA datasets, where Scan2Cap metrics are restricted to instances with $\geq 0.5$ IoU (\textbf{@0.5}) except L-A. We follow common practice to use thresholded Accuracy@0.5 (\textbf{A@0.5}), F1-score@0.5 (\textbf{F1@0.5}), and mean intersection over union (mIoU) for ScanRefer \cite{chen2020scanrefer} and Multi3DRefer \cite{zhang2023multi3drefer}. 
\label{sec:metrics}

\paragraph{Baselines and implementation.}
We adopt four baseline models from an array of distinct architectures to apply PoseAlign modifications, including LL3DA \cite{chen2024ll3da}, LL3DA-SONATA, Chat-Scene \cite{huang2024chat}, and 3D-LLAVA \cite{deng20253d}.
LL3DA-SONATA, referred to as LL3DA-S, is a variant of LL3DA whose point cloud encoder is switched to SONATA\cite{wu2025sonata}. 
All modifications belong to PoseAlign-Transform (\textit{PoseAlign-T}) except for Chat-Scene, which employs PoseAlign-Embed (\textit{PoseAlign-E}) due to its reliance on precomputed point cloud embeddings.

\begin{table*}[t]
\centering
\setlength{\tabcolsep}{4pt}
\setlength{\aboverulesep}{0pt} % Remove space above rules
\setlength{\belowrulesep}{0pt} % Remove space below rules
\renewcommand{\arraystretch}{1.05}
\resizebox{\linewidth}{!}{
\begin{tabular}{l|c|cc|cc|ccccG|ccccG}
\toprule
\toprule
\multirow{2}{*}{Model} & \multirow{2}{*}{Modality} &
\multicolumn{2}{c|}{ScanRef (val)} &
\multicolumn{2}{c|}{Multi3DRef (val)} &
\multicolumn{5}{c|}{ScanQA (val)} &
\multicolumn{5}{c}{Scan2Cap (val)} \\ 

\cmidrule(lr){3-16} 

&&A@0.5 & mIoU &
F1@0.5 & mIoU & 
C & B-4 & M & R & L-A &
% EM & EM-R & L-A & 
C@0.5 & B-4@0.5 & M@0.5 & R@0.5 & L-A\\ 
\midrule
\textit{\textbf{Specialist Models:}} &  &  &  &  & &  &  &  &  & &  &  &  &  &  \\

ScanQA \cite{azuma2022scanqa}          & PC&-	&-	&-	&-	&64.9	&10.1	&13.1	&33.3	&-	&-	&-	&-	&-	&-	\\
3D-VLP \cite{jin2023context}          & PC&-	&-	&-	&-	&67.0	&11.2	&13.5	&34.5	&-	&54.9	&32.3	&24.8	&51.5	&-	\\
3D-VisTA \cite{zhu20233d}        & PC&-	&-	&-	&-	&69.6	&10.4	&13.9	&45.7	&-	&61.6	&34.1	&26.8	&55.0	&-	\\
Scan2Cap \cite{chen2021scan2cap}        & PC&-	&-	&-	&-	&-	&-	&-	&-	&-	&39.1	&23.3	&22.0	&44.8	&-	\\
MORE \cite{jiao2022more}            & PC&-	&-	&-	&-	&-	&-	&-	&-	&-	&40.9	&22.9	&21.7	&44.4	&-	\\
SpaCap3D \cite{wang2022spatiality}       & PC&-	&-	&-	&-	&-	&-	&-	&-	&-	&44.0	&25.3	&22.3	&45.4	&-	\\
D3Net \cite{chen2022d}          & PC&-	&-	&-	&-	&-	&-	&-	&-	&-	&46.1	&30.3	&24.4	&51.7	&-	\\
UniT3D \cite{chen2023unit3d}         & PC&-	&-	&-	&-	&-	&-	&-	&-	&-	&46.7	&27.2	&21.9	&46.0	&-	\\
3DJCG \cite{cai20223djcg}          & PC&-	&-	&-	&-	&-	&-	&-	&-	&-	&49.5	&31.0	&24.2	&50.8	&-	\\
Vote2Cap-DETR \cite{chen2023end}  & PC&-	&-	&-	&-	&-	&-	&-	&-	&-	&61.8	&34.5	&26.2	&54.4	&-	\\
TGNN \cite{huang2021text}           & PC&-	&27.8	&-	&-	&-	&-	&-	&-	&-	&-	&-	&-	&-	&-	\\
M3DRef-CLIP \cite{zhang2023multi3drefer}    & PC&-	&35.7	&-	&32.6	&-	&-	&-	&-	&-	&-	&-	&-	&-	&-	\\
X-RefSeg3D \cite{qian2024x}     & PC&-	&29.9	&-	&-	&-	&-	&-	&-	&-	&-	&-	&-	&-	&-	\\
3D-STMN \cite{wu20243d}        & PC&-	&39.5	&-	&-	&-	&-	&-	&-	&-	&-	&-	&-	&-	&-	\\

\midrule
\textit{\textbf{Finetuned Generalists:}} &  &  &  &  & &  &  &  &  & &  &  &  &  & \\
3D-LLM \cite{hong20233d}     & PC+I &-	&-	&-	&-	&69.4	&12.0	&14.5	&35.7	&-	&-	&-	&-	&-	&-	\\
Scene-LLM \cite{fu2024scene}  & PC+I &-	&-	&-	&-	&80.0	&12.0	&16.8	&40.0	&-	&-	&-	&-	&-	&-	\\
SegPoint \cite{he2024segpoint}   & PC &-	&41.7	&-	&36.1	&-	&-	&-	&-	&-	&-	&-	&-	&-	&-	\\

\midrule
\textit{\textbf{Generalists:}} &  &  &  &  & &  &  &  &  & &  &  &  &  & \\
LEO \cite{huang2023embodied}            & PC&-	&-	&-	&-	&\textcolor{gray}{101.4}	&\textcolor{gray}{13.2}	&\textcolor{gray}{20.0}	&\textcolor{gray}{49.2}	&-	&72.4	&\textbf{38.2}	&\textbf{27.9}	&\textbf{58.1}	&-	\\
Scene-LLM \cite{fu2024scene}      & PC&-	&-	&-	&-	&80.0	&11.7	&15.8	&35.9	&-	&-	&-	&-	&-	&-	\\
Grounded 3D-LLM \cite{chen2024grounded} & PC&-	&-	&-	&-	&72.7	&13.4	&-	&-	&-	&70.6	&35.5	&-	&-	&-	\\

\midrule
\textit{\textbf{PoseRecover Benchmark:}} &  &  &  &  & &  &  &  &  & &  &  &  &  & \\
 LL3DA \cite{chen2024ll3da}
 & PC & - & - & - & - & 75.3 & 13.1 & 15.2 & 36.7 & 34.7 & 61.6 & 35.3 & 25.6 & 54.3 & 16.0 \\
 LL3DA + PoseAlign-T
 & PC & - & - &  -& - & 76.7 & 13.9 & 15.6 & 37.1 & 35.6 &\multicolumn{5}{c}{E: Rotated box} \\
% \midrule
LL3DA-S \cite{chen2024ll3da,wu2025sonata}
 & PC & - & - & - & - & 75.0 & 12.6 & 15.1 & 37.0 & 34.8 &\multicolumn{5}{c}{\multirow{2}{*}{E: No detector}} \\
LL3DA-S + PoseAlign-T
 & PC & - & - & - & - & 76.5 & 13.2 & 15.6 & 36.8 & 35.9  & \multirow{4}{*}{\ }\\
% \midrule
 Chat-Scene \cite{huang2024chat}
 & PC+I & 46.4 & - & \underline{49.1} & - & 85.6 & 15.6 & 17.8 & 40.4 & 43.9 
 % & 52.8 & 56.1 & 56.9 & 
 & 74.0 & 34.5 & 26.8 & 56.4 & 26.6\\
 Chat-Scene + PoseAlign-E
 & PC+I & \underline{46.9} & - & \textbf{50.2} & - & 87.2 & 15.1 & 18.1 & 41.1 & 44.7 
 % & 53.3 & 56.1 & 57.1 
 & 75.5 & 35.0 & \underline{27.1} & 56.7 & 26.9 \\
% \midrule
 3D-LLAVA \cite{deng20253d}
 & PC & 41.5 & \underline{42.6} & - & \underline{48.1} & \underline{95.4} & \underline{16.3} & \underline{18.9} & \underline{44.6} & \underline{45.7} 
 % & 54.9 & 57.3 & 57.9 
 & \textbf{77.4} & 36.4 & 26.9 & 57.4 & \underline{28.1} \\
 3D-LLAVA + PoseAlign-T
 & PC & \textbf{58.7} & \textbf{55.4} & - & \textbf{54.3} & \textbf{99.8} & \textbf{17.3} & \textbf{19.7} & \textbf{46.5} & \textbf{47.3} 
 % & 54.2 & 56.7 & 57.4 
 & \underline{76.1} & \underline{37.1} & \underline{27.1} & \underline{57.6} & \textbf{31.4} \\
\bottomrule
\bottomrule
\end{tabular}
}
\vspace{-0.3cm}
\caption{\textbf{Cross-dataset performance comparison} on multiple 3D vision-language tasks. `PC' and `I' represent point cloud and image modalities, respectively. 
Major metrics are highlighted with \colorbox{gray!20}{gray} background.
% LL3DA-S is our variant of LL3DA with SONATA as the point cloud encoder. 
% All our modifications belong to PoseAlign-Transform except for Chat-Scene which employs PoseAlign-Embed due to its fixture to precomputed point cloud embeddings. 
Performance on PoseRecover benchmark may differ from those in the original papers due to retraining with lower batch sizes. Baselines in PoseRecover benchmark are comparable with all methods because they do not use pose information, while our modifications are comparable within the benchmark due to additional pose input.}
\label{tab:posealign_results}
\vspace{-0.5cm}
\end{table*}

\paragraph{Training specifications.} All comparison methods are trained with only the instruction-tuning stage using respective original training schemes and datasets. For Chat-Scene \cite{huang2024chat} and 3D-LLAVA \cite{deng20253d}, the LLM LoRA \cite{hu2022lora} and the projection layer are trained while the 3D encoder is frozen. For LL3DA \cite{chen2024ll3da}, only the Q-Former \cite{li2023blip} is trained. Freezing the encoder prevents the formation of preference over objects on the positive x-axis, ensuring fairness in segmentation- or detection-based assessments. All point cloud data augmentations have been disabled for PoseAlign variants to precisely align the egocentric point cloud coordinates, which include random rotation, flipping, jittering, and scaling. We deem that data augmentation is compensated by the randomness of ego poses during pose selection, where PoseAlign-Transform and -Embed can be seen as exotic implementations of rotation-shift augmentation.
\label{sec:training_specifications}

\subsection{Main Results}
In this experiment, we validate the effectiveness of PoseRecover and PoseAlign designs with an array of specialist and generalist 3D LMMs on the ScanRefer, Multi3DRefer, ScanQA, and Scan2Cap datasets in Table \ref{tab:posealign_results}. All modified models on the PoseRecover Benchmark get the additional pose input retrieved using the PoseRecover script. The performance of LEO \cite{huang2023embodied} is colored in gray as it utilizes the ground truth object features, which forms a slightly stronger setting as discussed in section \ref{sec:difference_with_leo}. LL3DA has poor compatibility with rotated boxes, and SONATA is a feature extractor with no detector component, so some results on Scan2Cap are omitted. Methods enhanced with PoseAlign receive performance boosts broadly across datasets and metrics, where 3D-LLAVA modified with PoseAlign-Transform hits the highest overall performance of $55.4\%$\textcolor{cvprblue}{\footnotesize$(\Delta30.0\%)$} ScanRefer mIoU, $54.3\%$\textcolor{cvprblue}{\footnotesize$(\Delta12.9\%)$} Multi3DRefer mIoU, $47.3\%$\textcolor{cvprblue}{\footnotesize$(\Delta3.5\%)$} and $31.4\%$\textcolor{cvprblue}{\footnotesize$(\Delta11.7\%)$} LLM-as-judge accuracy on ScanQA and Scan2Cap, respectively. We conclude that PoseAlign unleashes the full capabilities of current 3D LMMs, which were once bounded by the ill-posed problem definition in existing benchmarks.

\paragraph{Is PoseAlign a good addition to existing tasks?} We focus on the PoseRecover benchmark to validate the effectiveness of PoseAlign. Comparing all four baselines with our respective modifications, performance boosts can be widely seen across QA datasets with an average bump of $1\%$ and $0.6\%$ on ScanQA and Scan2Cap, respectively, including CiDEr, BLEU-4, METEOR, and ROUGE-L on all four backbones. The LLM-as-judge accuracy boosts are greater for PoseAlign-T variants ($35.6\%$\textcolor{cvprblue}{\footnotesize$(\Delta2.6\%)$} on LL3DA, $35.9\%$\textcolor{cvprblue}{\footnotesize$(\Delta3.2\%)$} on LL3DA-S, $47.3\%$\textcolor{cvprblue}{\footnotesize$(\Delta3.5\%)$} on 3D-LLAVA) than PoseAlign-E variants ($44.7\%$\textcolor{cvprblue}{\footnotesize$(\Delta1.8\%)$} on Chat-Scene), signaling a better realization of direction awareness. On refer segmentation benchmarks, both Chat-Scene and 3D-LLAVA receive performance boosts among which 3D-LLAVA receives the largest $+12.8\%$ and $+6.2\%$ mIoU on ScanRefer and Multi3DRefer, respectively. Such performance is astounding given that the point cloud segmentor (\textit{i.e.,} 3D encoder) is frozen, which attributes all improvements to improved `$<$SEG$>$' token quality generated by the LLM \cite{deng20253d}. We conclude that, given their potential to improve performance, ego-poses are undoubtedly good additions to current 3D LMM pipelines.
\vspace{-0.3cm}

\paragraph{Is LLM-as-judge a good metric?} 
As discussed in section \ref{sec:metrics}, a good language-task metric should be consistent under verbal randomness as confirmed by Table \ref{tab:posealign_results}. LLM-as-judge accuracy is consistently higher across all our modifications than the respective baselines, while other metrics tend to fluctuate. 
It is also robust to the judgment model, as shown in Tables \ref{tab:posealign_results} and \ref{tab:gpt_5_results}. We conclude that LLM-as-judge is a sufficiently good metric for this task.

\begin{table}[t]
\centering
\setlength{\tabcolsep}{3pt}
\renewcommand{\arraystretch}{1.05}
\resizebox{\linewidth}{!}{
\begin{tabular}{l|cc|cc}
\toprule
\multirow{2}{*}{Design} &
ScanRef &
Multi3DRef &
ScanQA &
Scan2Cap \\ 
% \cmidrule(lr){2-6} 

& mIoU$\uparrow$ & mIoU$\uparrow$ & L-A$\uparrow$ &
L-A$\uparrow$\\ 
\midrule
Baseline &42.6	&48.1	&45.7	&28.1 \\
Baseline PoseAlign-T(Top)&37.5    &41.5   &43.0   &23.5 \\
Random Pose & 39.0	&44.3 &44.7  &25.8\\
PoseAlign-T(Clip X=0.3)&\underline{55.4}	&\underline{54.3}	&\underline{47.3}	&\textbf{31.4} \\
PoseAlign-T(Top)&\textbf{68.5}	&\textbf{60.2}	&\textbf{47.5}	&\underline{29.7} \\
PoseAlign-E(Top)&43.2 	&49.3   &43.4	&23.5 \\
PoseAlign-P(Top)&44.2	&49.4	&44.9	&28.1 \\
\bottomrule
\end{tabular}
}
\vspace{-0.2cm}
\caption{\textbf{Ablation experiment on 3D-LLAVA}. `Baseline PoseAlign-T' is the performance of the baseline model on PoseAlign-T data, where the input point cloud is transformed to the camera location following Equation \ref{eq:posealign_transform}. `Random Pose' uses random camera poses instead of those found by PoseRecover.}
\label{tab:ablation}
\vspace{-0.4cm}
\end{table}

\subsection{Other Results}
\paragraph{Ablation on 3D-LLAVA.} We compare the effect of different PoseAlign designs on 3D-LLAVA in Table \ref{tab:ablation}. Incorporating ego-pose information clearly benefits spatial reasoning, as the proposed PoseAlign-Transform consistently outperforms all other variants on refer segmentation tasks, improving mIoU by up to $68.5\%$\textcolor{cvprblue}{\footnotesize$(\Delta60.8\%)$} on ScanRefer and $60.2\%$\textcolor{cvprblue}{\footnotesize$(\Delta25.2\%)$} on Multi3DRefer. The Clip strategy ($X = 0.3$) for PoseAlign-Transform further stabilizes performance metrics across benchmarks by purging inconsistent views. In contrast, encoding poses as text (PoseAlign-Prompt) or projection features (PoseAlign-Embed) offers limited or inconsistent gains. Additionally, test performance of a well-trained 3D-LLAVA baseline model on PoseAlign-Transform data (Baseline PoseAlign-T) significantly degrades all metrics, which gives two important insights: 1) the segmentation model never uses the pose data as a shortcut because it is frozen, and 2) the current performance increments are brought purely by better `$<$SEG$>$' tokens. Using random poses also decreases model performance slightly below baseline, showcasing the importance of PoseRecover design. We conclude that PoseAlign-Transform with the Clip strategy where $X=0.3$ is the most balanced and robust way of introducing directional awareness into 3D LMMs.
% PoseAlign-Transform with Clip Ratio $X=0.3$ is selected as the default model for all other experiments if not otherwise specified.

\begin{table}[t]
\centering
\setlength{\tabcolsep}{3pt}
\renewcommand{\arraystretch}{1.05}
\resizebox{0.8\linewidth}{!}{
\begin{tabular}{l|cccc}
\toprule
&
ScanRef &
Multi3DRef &
ScanQA &
Scan2Cap \\ 
% \cmidrule(lr){2-6} 

$X$ & mIoU$\uparrow$ & mIoU$\uparrow$ & L-A$\uparrow$ &
L-A$\uparrow$\\ 
\midrule
0.0     &55.1	&\underline{54.2}	&\textbf{47.3}	&30.1\\
0.1     &55.2	&\textbf{54.3}	&46.6	&30.8\\
0.2     &54.8	&\textbf{54.3}	&\textbf{47.3}	&\underline{31.2}\\
0.3     &\textbf{55.4}	&\textbf{54.3}	&\textbf{47.3}	&\textbf{31.4}\\
0.4     &\textbf{55.4}	&\textbf{54.3}	&\underline{46.9}	&30.9\\
0.45    &\underline{55.2}	&\textbf{54.3}	&46.5		&\textbf{31.4}\\
0.49    &55.1	&53.9	&\underline{46.9}		&31.1\\
\bottomrule
\end{tabular}
}
\vspace{-0.2cm}
\caption{\textbf{Parameter tuning experiment} for Clip Ratio X of PoseAlign-Transform on 3D-LLAVA.}
\label{tab:parameter_tuning}
\vspace{-0.4cm}
\end{table}

\paragraph{Parameter tuning.} Table \ref{tab:parameter_tuning} analyzes the sensitivity of the Clip Ratio $X$ in PoseAlign-Transform, which discards the top and bottom $X\%$ of views based on object–frustum intersection during selection of the mission-critical pose. Despite theoretical concerns about opposite-view errors discussed in Figure \ref{fig:clip_yaw}, the model performance remains stable across datasets for $0.0 \leq X \leq 0.49$ with a slight hill in the mediocre Clip Ratio ranging from $0.2 \leq X \leq 0.4$, demonstrating the robustness of the method to this hyperparameter. This is because a smaller variance in viewing directions indeed enhances pose coherence but destroys data diversity, while higher variance in viewing directions adds noise to the model but also serves as effective data augmentation as discussed in section \ref{sec:training_specifications}. Moderate clipping (around $X=0.3$) yields the best balance, slightly improving results on ScanQA and Scan2Cap to $47.3\%$ and $31.4\%$, respectively. Therefore, $X=0.3$ is adopted as the default value for all other experiments.

\begin{table*}[t]
\centering
\small
\setlength{\tabcolsep}{3pt}
\renewcommand{\arraystretch}{1.05}
\resizebox{\linewidth}{!}{
\begin{tabular}{l|c|ccccG|ccccG}
\toprule
\multirow{2}{*}{Model}&\multirow{2}{*}{Evaluated on}&\multicolumn{5}{c|}{ScanQA ($46.7\%$ Direction-critical)} & \multicolumn{5}{c}{Scan2Cap ($89.7\%$ Direction-critical)}\\
\cmidrule{3-12}
&& C$\uparrow$ & B-4$\uparrow$ & M$\uparrow$ & R$\uparrow$ & L-A$\uparrow$ &
C$\uparrow$ & B-4$\uparrow$ & M$\uparrow$ & R$\uparrow$ & L-A$\uparrow$\\ 
% \cmidrule(lr){2-6} 
\midrule
\multirow{3}{*}{3D-LLAVA} & Full Val set
&95.4	&16.3	&18.9	&44.6	&45.7    &77.4	&36.4	&26.93	&57.4	&28.1\\
& Direction-critical subset
&91.6	&18.0	&17.8	&41.0	&37.8    &76.8	&36.8	&26.94	&57.4	&28.0\\
& Complementary subset
&97.6	&11.9	&20.1	&47.7	&52.4    &82.7	&32.6	&26.88	&56.9	&29.6\\
\midrule
\multirow{3}{*}{\begin{tabular}{@{}c@{}}3D-LLAVA+\\ PoseAlign-T\end{tabular}} & Full Val set& 99.8\textcolor{cvprblue}{(+4.4)} &	17.3\textcolor{cvprblue}{(+1)} &	19.7\textcolor{cvprblue}{(+0.8)} &	46.5\textcolor{cvprblue}{(+1.9)} &	47.3\textcolor{cvprblue}{(+1.6)} &				76.1\textcolor{cvprblue}{(-1.3)} &	37.1\textcolor{cvprblue}{(+0.7)} &	27.11\textcolor{cvprblue}{(+0.18)} &	57.6\textcolor{cvprblue}{(+0.2)} &	31.4\textcolor{cvprblue}{(+3.3)}	\\
& Direction-critical subset& 96.1\textcolor{cvprblue}{(+4.5)} &	19.6\textcolor{cvprblue}{(+1.6)} &	18.7\textcolor{cvprblue}{(+0.9)} &	43.1\textcolor{cvprblue}{(+2.1)} &	40.3\textcolor{cvprblue}{(+2.5)} &				75.9\textcolor{cvprblue}{(-0.9)} &	37.5\textcolor{cvprblue}{(+0.7)} &	27.12\textcolor{cvprblue}{(+0.18)} &	57.7\textcolor{cvprblue}{(+0.3)} &	31.3\textcolor{cvprblue}{(+3.3)}	\\
& Complementary subset& 101.8\textcolor{cvprblue}{(+4.2)} &	11.5\textcolor{cvprblue}{(-0.4)} &	20.8\textcolor{cvprblue}{(+0.7)} &	49.4\textcolor{cvprblue}{(+1.7)} &	53.3\textcolor{cvprblue}{(+0.9)} &				77.8\textcolor{cvprblue}{(-4.9)} &	33.4\textcolor{cvprblue}{(+0.8)} &	27.10\textcolor{cvprblue}{(+0.22)} &	56.9\textcolor{cvprblue}{(+0.0)} &	32.6\textcolor{cvprblue}{(+3.0)}	\\

\bottomrule
\end{tabular}
}
\vspace{-0.2cm}
\caption{Performance on direction-critical question subset.}
\label{tab:direction_critical_subset}
\end{table*}

\begin{figure*}[t]
  \centering
   \vspace{-0.3cm}
   \includegraphics[width=\linewidth]{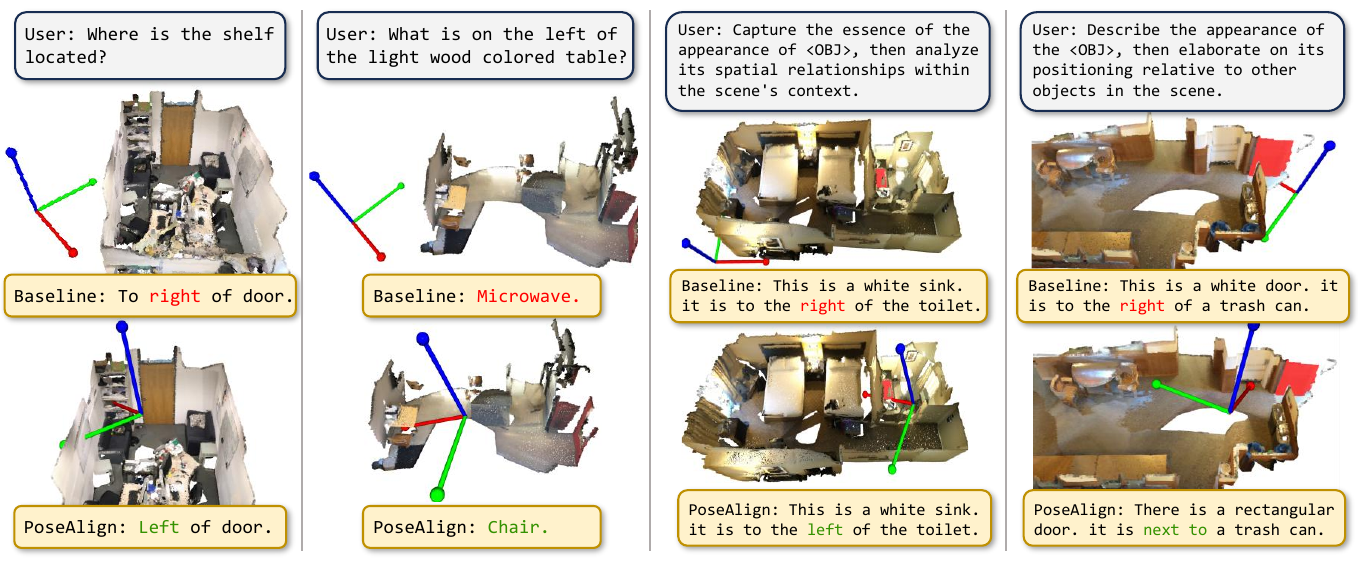}
   \vspace{-0.7cm}
   \caption{\textbf{Qualitative results of direction-critical questions} for 3D-LLAVA baseline (top row) and PoseAlign-Transform (bottom row). The XYZ axes of the world coordinate frame are colored with \textcolor{red!70!black}{red}, \textcolor{green!50!black}{green}, and \textcolor{blue!80!black}{blue}, respectively. The baseline paradigm uses default world coordinates of ScanNet-v2, which are non-informative. Instead, the PoseAlign paradigm aligns the coordinate frame to the recovered ego pose, providing an anchor for robust spatial reasoning.
   % Baselines are \textcolor{red!70!black}{incorrect} because the default world coordinate only adds confusion}
   \textcolor{red!70!black}{Red text} highlights wrong answers and \textcolor{green!50!black}{green text} highlights correct answers.}
   \label{fig:visualization}
   \vspace{-0.4cm}
\end{figure*}

\paragraph{Performance on direction-critical subsets.} Table \ref{tab:direction_critical_subset} compares the baseline 3D-LLAVA and its pose-aware variant (PoseAlign-Transform) on full validation sets, direction-critical subsets, and their respective complementary direction-agnostic subsets of ScanQA and Scan2Cap. Firstly, the direction-critical subsets identified via LLMs exhibit a notable performance drop in the baseline model, confirming that existing 3D LMMs struggle with questions requiring spatial orientation. Applying PoseAlign-Transform consistently improves all metrics, especially the LLM-as-judge accuracy, narrowing down the gap between the two subsets. The gains are especially pronounced on ScanQA, where PoseAlign-T achieves $96.1$\textcolor{cvprblue}{\footnotesize$(\Delta4.9\%)$} CiDEr and $40.3\%$\textcolor{cvprblue}{\footnotesize$(\Delta6.6\%)$} LLM-as-judge accuracy. While the direction-critical subset is the best-effort judgment of a powerful LLM, some direction-critical questions may still leak into the complementary subset, causing performance to rise slightly there as well. We conclude that our PoseAlign design indeed enhances directional reasoning based on the substantial improvements, especially on the direction-critical subsets.

\subsection{Qualitative Results}
As displayed in Figure \ref{fig:visualization}, baseline models struggle to perceive spatial relationships, where the default world coordinate only adds confusion and complicates spatial reasoning. In contrast, PoseAlign effectively avoids directional ambiguity thanks to the recovered ego pose, enabling robust and consistent spatial reasoning for 3D LMMs.

\subsection{Limitations} 
Despite being lightweight and generic, our method has several limitations, including assuming SLAM accuracy, the variety of training views, and a static environment. We kindly refer readers to section \ref{sec:limitations} for a detailed discussion.

%% file: sec/6_conclusion.tex
\section{Conclusion}
\label{sec:conclusion}
We have proposed PoseRecover and PoseAlign, a pair of lightweight yet powerful techniques that transform existing 3D LMMs towards a rigorously defined direction-aware paradigm. To fix existing benchmarks, PoseRecover automatically reconstructs mission-critical ego poses by aligning object annotations with camera frustums derived from ScanNet RGB-D extrinsics, effectively correcting ill-posed direction-critical queries across existing datasets. Building upon these recovered poses, PoseAlign enables persistent directional awareness for 3D LMMs by transforming the input point cloud or encoded point cloud embeddings into the camera reference frame, thereby resolving ambiguity in ego direction and substantially improving spatial reasoning performance. Extensive experiments on multiple benchmarks and model architectures prove that PoseRecover and PoseAlign together unlock the latent potential of current 3D LMM architectures, achieving consistent performance gains without retraining the point cloud encoder. We advocate this framework as a simple, generic, and effective paradigm for advancing direction-aware 3D-language understanding of LMMs.

%% file: sec/acknowledgment.tex
% \vspace{-0.2cm}
\section*{Acknowledgments}
% \vspace{-0.2cm}
This research is supported by Networked Exchange, United Strength for Stronger Partnerships between Japan and ASEAN (NEXUS), a collaboration program between the Agency for Science, Technology and Research (A*STAR), Singapore (Grant No. R2416IR138), and the Japan Science and Technology Agency (JST), Japan (Grant No. JPMJNX25CA). Weihao Xuan is supported by the RIKEN Junior Research Associate (JRA) Program.\\

%% file: sec/7_appendix_v2.tex
\clearpage
\setcounter{page}{1}
\maketitlesupplementary

\appendix

\section{Additional Notes}

\subsection{Explanations}

\paragraph{Ill-posed questions without poses.} At first glance, it may seem odd that questions involving left/right directions are ill-posed when ego pose is not present, especially to researchers in the fields of image or VLM research. A key distinction between images and 3D point clouds is that images are captured from an ego pose, which inherently defines left/right as the image left/right; However, a room-level point cloud is reconstructed using RGB-D video or even with 360 degree view LiDARs. Once the points are put into free 3D space, the concept of an ego reference frame disappears, now that the ego pose can no longer be trivially inferred from the room-level point cloud data itself. However, existing 3D LMMs receive only such a room-level point cloud and are asked to answer questions containing directional words, which is an ill-posed problem we aim to conquer in this paper. 
\paragraph{From implicit ego pose inference to explicit ego pose input.} We could infer from past research that they want the 3D LMM to implicitly infer its ego position first, then derive the related answers based on that inferred ego pose. However, this regime is both (1) inherently ill-posed, and (2) adds unnecessary complexity for the 3D LMM to infer its ego position, while the robot's ego pose is trivially available in realistic applications such as SLAM-capable embodied intelligence operating indoors. We therefore conclude that adding such pose data as additional input to the 3D LMMs does not change its application potential at all, and such conversion of regime is both practical and theoretically grounded.

\paragraph{From an ego pose trajectory to a mission-critical pose.} In practice, even when the ego trajectory is a free lunch, the agent does not always face the target object of the query and the mission-critical pose has to be assessed on-the-fly. However, we argue that such a mission-critical pose should be retrieved with other techniques rather than with the 3D LMM itself, as the task is essentially a cross-modal alignment problem with well-established solutions. For example, the agent can localize the mission-critical area by comparing the text embedding of the queried object with an offline precomputed semantic map of the indoor point cloud, e.g., using CLIP \cite{radford2021learning} or OpenScene \cite{peng2023openscene}. After that, using our PoseRecover script, the agent can locate the most probable ego position from the ego pose trajectory, which can then be used to reason consistently and unambiguously. In this work, we are more interested in the prior problem of \textit{whether a mission-critical pose retrieved from ground truth is a necessary and beneficial addition to 3D LMM input, to rectify the ill-posed direction definitions}, rather than a full end-to-end implementation; Our work is purely exploratory and there must exist some better way of integrating those poses into a coherent pipeline than with our PoseAlign.
\label{sec:annotation_acquisition}

\subsection{Discussions}

\paragraph{Compatibility with image-based methods.} We deem all image-based methods to have implicitly used ego pose trajectory during the camera projection of the input images. Therefore, these image-based methods are already implicit implementations of the PoseAlign-Transform method and do not need further modifications.

\paragraph{Compatibility with SQA3D.} Notably, PoseAlign receives negative results on SQA3D as listed in Tables \ref{tab:parameter_tuning_chat_scene}-\ref{tab:parameter_tuning_3d_llava}. This is because among all training datasets, SQA3D is the only dataset that does not provide any object annotations, which means that PoseRecover pipeline cannot determine authentic poses for the questions, yielding random poses instead. This results in a slightly decreased performance compared to the baseline. We list supplementing object annotations and poses to the SQA3D benchmark as a major future work.

\paragraph{Difference compared to the setting of LEO \cite{huang2023embodied}.} While both PoseRecover and LEO can utilize the ground truth object label, we want to point out that they are different settings. PoseRecover aims to rule out the directional ambiguity problem which requires a pose that can only be retrieved using object annotations. PoseAlign only feeds a pose facing the object into the point cloud encoder and the point cloud encoder is not tuned at all, which forbids formation of any direct object feature shortcuts like presented in LEO. We are fully aware that the performance increment of PoseAlign is brought by the normally unavailable pose provided by PoseRecover, but we see this as a testament to our hypothesis that a mission-critical pose derived prior to 3D LMMs is a great tool for ruling out reasoning ambiguity. The system in real action should be composed of two stages, where the agent first `imagines' itself to be at the mission location and then derives the answers based on that reference frame. 
% This is a better way than delegating both pose retrieval and spatial reasoning to the 3D LMM, because the former problem is fundamentally not suitable for 3D LMM \cite{ma2022sqa3d,zhang2025point}. 
Furthermore, the way PoseRecover generates the pose is non-deterministic and the target pose is randomly chosen from all poses that can see the target object, which makes it a lot more realistic and forgiving than LEO setup as shown in Table \ref{tab:posealign_results}. More explanation can be found in section \ref{sec:annotation_acquisition}.
\label{sec:difference_with_leo}

\begin{table}[t]
\centering
\resizebox{\linewidth}{!}{
\begin{tabular}{l|c|cccc}
\toprule
&&ScanRefer &  Multi3DRef & ScanQA & Scan2Cap\\
Model & Noise Bound & mIoU$\uparrow$ & mIoU$\uparrow$ & B-4$\uparrow$ & B-4$\uparrow$ \\
% Model & Noise Bound & ScanRef mIoU$\uparrow$ & Multi3DRef mIoU$\uparrow$ & ScanQA B-4$\uparrow$ & Scan2Cap B-4$\uparrow$\\ 
\midrule
3D-LLAVA & / & 42.6	& 48.1	& 16.3	&36.4\\
\midrule
\multirow{3}{*}{\begin{tabular}{@{}c@{}}3D-LLAVA+\\ PoseAlign-T\\ (Top)\end{tabular}}& 0\degree, 0m & 68.5	& 60.2	& 17.3	&37.1\\
&15\degree, 0.2m    &62.5	&57.7	&17.1	&36.9\\
&30\degree, 0.5m    &53.2	&55.1	&17.0	&36.6\\
\bottomrule
\end{tabular}
}
\vspace{-0.2cm}
\caption{Robustness to added uniform pose error.}
\label{tab:rebuttal_pose_error}
\vspace{-0.3cm}
\end{table}

\begin{figure}[t]
  \centering
   \includegraphics[width=0.9\linewidth]{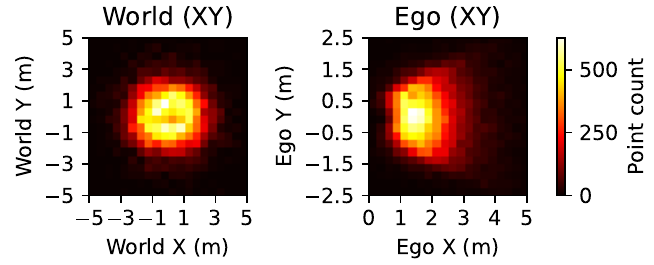}
   \vspace{-0.3cm}
   \caption{Distribution of objects in world (L) and PoseAlign ego (R) coordinates on X-Y plane. While clustered in front view, objects still follow non-trivial distribution.}
   \label{fig:rebuttal_heatmap}
   \vspace{-0.3cm}
\end{figure}

\paragraph{Applicability to RGB-D data without camera extrinsics or LiDAR data.}
In practical embodied workflows, camera extrinsics are a free lunch from SLAM. For datasets lacking extrinsics, they can be recovered offline using ORB-SLAM2 or similar methods. 
We simulated SLAM pose noises in Table \ref{tab:rebuttal_pose_error} which reveals the robustness of our method.
For single-frame RGB-D or LiDAR data without extrinsics, which are too sparse for SLAM, we operate on single-frame point clouds, which naturally lie in the ego frame and thus constitute a natural application of our method.

\paragraph{Does our setup introduce information leakage?} 
First, we clarify that our benchmark is intentionally defined as an ego-conditioned understanding task, rather than standard free-view understanding. As such, ego alignment is part of the task specification rather than an unintended shortcut.
Second, to reduce unintended bias, we avoid selecting the single best-matching viewpoint and instead sample from a diverse candidate set, resulting in non-trivial object spatial distributions in the horizontal plane (see Figure\ \ref{fig:rebuttal_heatmap}).
Third, we freeze the point cloud encoder to prevent learning geometric shortcuts, further limiting overfitting to pose priors.
Fourth, 
% full directional awareness means both `being in the room' and `facing the target'. A
a PoseAlign-trained 3D-LLAVA model tested on random camera extrinsics results in ScanRefer mIoU dropping from 55.4\% to 42.9\% but still above vanilla 42.6\%. So target facing is crucial to directional awareness but is not a shortcut, otherwise the performance would drop below vanilla.

\paragraph{Interpretation of the performance superiority of PoseAlign-T.} The pretrained 3D encoder is trained on cartesian coordinates which internalizes a concept of distance and directions based on the input coordinates. Compared to PoseAlign-T which guarantees consistency at input, PoseAlign-E and PoseAlign-P variants cannot guarantee alignment with (or even fight against) the encoder's concept of direction, causing deterioration.
\label{sec:appendix_transform_superiority_discussion}

\subsection{Limitations}
While PoseRecover and PoseAlign provide a lightweight and general solution for introducing directional awareness into existing 3D-LMMs, several limitations remain. First, our method relies on the assumption that the 3D LMM is applied on embodied agents, where the availability and accuracy of camera extrinsics are guaranteed. Noise from inaccurate sensor calibration or synchronization is not considered. Second, the recovered poses are limited to the discrete set of views present in ScanNet and may not fully represent all possible viewpoints implied by the language queries, e.g., `standing on the kitchen counter' or `crouching under the desk'. Third, although PoseAlign improves directional reasoning without retraining the point cloud encoder, it still assumes the encoder’s coordinate sensitivity is preserved, which may not hold for certain architectures employing rotation-invariant designs. Finally, our approach focuses on static indoor environments; extending it to dynamic or outdoor scenarios where ego motion and scene layout change continuously remains an open direction.
\label{sec:limitations}

\begin{table}[t]
\centering
\setlength{\tabcolsep}{3pt}
\renewcommand{\arraystretch}{1.05}
\resizebox{0.85\linewidth}{!}{
\begin{tabular}{l|ccc}
\toprule
\multirow{2}{*}{Design} &
ScanQA &
SQA3D &
Scan2Cap \\ 
% \cmidrule(lr){2-6} 
& L-A$\uparrow$ & L-A$\uparrow$&
L-A$\uparrow$\\ 
\midrule
LL3DA                   &35.2&	-   &	23.3\\
LL3DA+PoseAlign-T       &35.4&	-   &	-   \\
LL3DA-S                 &36.3&	-   &	-   \\
LL3DA-S+PoseAlign-T     &36.5&	-   &	-   \\
Chat-Scene              &44.1&	56.9&	38.6\\
Chat-Scene+PoseAlign-E  &44.7&	56.8&	\underline{39.7}\\
3D-LLAVA                &\underline{45.9}&	\textbf{58.1}&	39.3\\
3D-LLAVA+PoseAlign-T    &\textbf{48.3}&	\underline{57.3}&	\textbf{40.7}\\
\bottomrule
\end{tabular}
}
\caption{Comparison between baselines and respective PoseAlign variants using GPT-5-mini for LLM-as-judge.}
\label{tab:gpt_5_results}
\vspace{-0.3cm}
\end{table}

\begin{table*}[t]
\centering
\setlength{\tabcolsep}{3pt}
\renewcommand{\arraystretch}{1.05}
\resizebox{0.7\linewidth}{!}{
\begin{tabular}{l|ccccG|ccG|ccccG}
\toprule
\multirow{2}{*}{$X$} &
\multicolumn{5}{c|}{ScanQA (val)} &
\multicolumn{3}{c|}{SQA3D (test)} &
\multicolumn{5}{c}{Scan2Cap (val)} \\ 
\cmidrule(lr){2-14} 
& C & B-4 & M & R & L-A &
EM & EM-R & L-A & 
C@0.5 & B-4@0.5 & M@0.5 & R@0.5 & L-A\\
\midrule
None    & 85.6	& 15.6	& 17.8	& 40.4	& 43.9	& 52.8	& 56.1	& 56.9	& 74.0	& 34.5	& 26.8	& 56.4	& 26.6	\\
0.0     & 83.8	& 14.2	& 17.4	& 39.7	& 43.1	& 53.4	& 56.2	& 57.2	& 37.2	& 27.9	& 23.4	& 52.0	& 9.76	\\
0.1     & 86.7	& 14.9	& 17.9	& 40.8	& 44.3	& 52.6	& 55.7	& 56.6	& 74.1	& 34.7	& 26.9	& 56.5	& 27.0	\\
0.2     & 87.2	& 15.1	& 18.1	& 41.1	& 44.7	& 53.3	& 56.1	& 57.1	& 75.5	& 35.0	& 27.1	& 56.7	& 26.9	\\
0.3     & 87.5	& 15.5	& 17.9	& 40.9	& 44.8	& 53.3	& 56.2	& 57.1	& 74.9	& 34.7	& 26.9	& 56.5	& 26.9	\\
0.4     & 82.7	& 14.3	& 17.0	& 39.2	& 42.7	& 52.8	& 55.4	& 56.1	& 27.5	& 25.6	& 22.4	& 50.6	& 6.2	\\
0.45    & 84.3	& 14.0	& 17.5	& 39.9	& 42.9	& 52.8	& 55.4	& 56.3	& 28.2	& 26.4	& 22.7	& 51.1	& 6.8	\\
0.49    & 83.6	& 15.4	& 17.4	& 39.5	& 42.8	& 52.8	& 55.5	& 56.3	& 60.7	& 32.2	& 25.8	& 54.9	& 19.0	\\
\bottomrule
\end{tabular}
}
\caption{\textbf{Full parameter tuning experiment} for Clip Ratio X of PoseAlign-Embed on Chat-Scene.}
\label{tab:parameter_tuning_chat_scene}
\end{table*}

\begin{table*}[t]
\centering
\setlength{\tabcolsep}{3pt}
\renewcommand{\arraystretch}{1.05}
\resizebox{0.7\linewidth}{!}{
\begin{tabular}{l|ccccG|ccG|ccccG}
\toprule
\multirow{2}{*}{$X$} &
\multicolumn{5}{c|}{ScanQA (val)} &
\multicolumn{3}{c|}{SQA3D (test)} &
\multicolumn{5}{c}{Scan2Cap (val)} \\ 
\cmidrule(lr){2-14} 
& C & B-4 & M & R & L-A &
EM & EM-R & L-A & 
C@0.5 & B-4@0.5 & M@0.5 & R@0.5 & L-A\\
\midrule
None    &95.4	&16.3	&18.9	&44.6	&45.7	&54.9	&57.3	&57.9	&77.4	&36.4	&26.9	&57.4	&28.1	\\
0.0     &98.7	&16.3	&19.3	&46.2	&47.3	&55.1	&57.6	&56.1	&75.5	&36.6	&26.9	&57.2	&30.1	\\
0.1     &99.4	&17.4	&19.4	&46.1	&46.6	&52.6	&55.0	&55.6	&76.6	&37.0	&27.1	&57.5	&30.8	\\
0.2     &99.8	&17.3	&19.5	&46.5	&47.3	&53.4	&56.2	&56.6	&74.7	&36.9	&27.0	&57.5	&31.2	\\
0.3     &99.8	&17.3	&19.7	&46.5	&47.3	&54.2	&56.7	&57.4	&76.1	&37.1	&27.1	&57.6	&31.4	\\
0.4     &96.9	&15.6	&19.0	&45.6	&46.9	&53.5	&56.1	&55.9	&73.6	&36.4	&26.8	&57.4	&30.9	\\
0.45    &97.8	&16.0	&19.2	&45.8	&46.5	&53.9	&56.0	&56.6	&75.7	&36.9	&27.0	&57.4	&31.4	\\
0.49    &98.2	&16.6	&19.3	&46.1	&46.9	&54.2	&56.7	&55.8	&76.3	&37.3	&27.1	&57.5	&31.1	\\
\bottomrule
\end{tabular}
}
\caption{\textbf{Full parameter tuning experiment} for Clip Ratio X of PoseAlign-Transform on 3D-LLAVA.}
\label{tab:parameter_tuning_3d_llava}
\end{table*}

\section{Additional Experiments}

\subsection{Other Experimental Setup}
\paragraph{Diversity of backbone architectures.} We adopt four baseline models, including LL3DA \cite{chen2024ll3da}, LL3DA-SONATA, Chat-Scene \cite{huang2024chat} and 3D-LLAVA \cite{deng20253d}. Their encoder backbones have distinct architectures ranging from BERT-style Vote2Cap-DETR operating on point patches \cite{chen2023end}, self-supervised PointTransformerV3 with serialized voxels \cite{wu2025sonata}, sparse convolutional neural network with sparse voxels \cite{schult2022mask3d}, and Omni Superpoint Transformer operating on superpoints \cite{deng20253d}. They include both object-wise \cite{chen2024ll3da,huang2024chat} and point-wise \cite{wu2025sonata,deng20253d} feature projection, both with or without a Q-Former \cite{dai2023instructblip}. Their base LLMs include OPT-1.3B, Vicuna-v1.5-7B, and the LLM section of LLAVA-v1.5-7B. Consistent performance improvements despite such diversity proves the universal applicability of the PoseAlign modification. 

\paragraph{Hardware specification.} All comparison methods and respective improved versions are trained on $4\times$ RTX 4090 24G. Batch sizes are trimmed to the same value within the same backbone to enable fair comparison.

\paragraph{PoseAlign-Embed implementation.} For the only backbone enhanced with PoseAlign-Embed, Chat-Scene has been a special case among all object-centric 3D LMMs because it passes both the location and the feature of an object into the projection module. In that sense, encoding the ego-pose feature in Chat-Scene reduces to transforming the object locations following Equation \ref{eq:posealign_transform}. For other backbones where the object location is not passed to the LLM, similar modifications can be done by adding a sinusoidal position embedding of the transformed object location onto the projection feature.

\paragraph{LLM-as-judge configuration.} For GPT-OSS-20B, we use the officially-recommended vllm package, which packs information into harmony format. The reasoning effort is set to medium in the system prompt. A stronger assessment suite is enabled with GPT-5-mini using minimal reasoning budget.

\subsection{Additional Results}

\paragraph{Direction-critical question distribution.} We apply the template in section \ref{sec:appendix_direction_gpt_prompt} on the validation sets of three existing point cloud language datasets, ScanQA, Scan2Cap, and SQA3D. The respective portions of direction-critical questions are 46.7\% on ScanQA, 89.7\% on Scan2Cap, and 95.2\% on SQA3D, respectively. The high ratio for the latter two datasets are due to their strong dependence on allocentric spatial relationships to define the caption target or the ego position. We conclude that a significant portion of existing 3D LMM benchmarks contain direction-critical questions which are ill-posed without ego poses. 
\label{sec:appendix_direction_critical_judgement}

\paragraph{Results for LLM-as-judge with GPT-5.}
In addition to reporting accuracy judged by GPT-OSS-20B, we also provide judgements from the powerful GPT-5 with the same set of prompts, which are listed in Table \ref{tab:gpt_5_results}. Similar trends are observed as found in Table \ref{tab:posealign_results}, where PoseAlign variants improve LLM-as-judge metrics consistently across all baselines and the majority of datasets. Notably, PoseAlign achieves $48.3\%$\textcolor{cvprblue}{\footnotesize$(\Delta5.0\%)$} and $40.7(+1.4)\%$\textcolor{cvprblue}{\footnotesize$(\Delta3.6\%)$} accuracy on ScanQA and Scan2Cap, and achieving top-tier accuracy of $57.3\%$ on SQA3D. We also note that the GPT-OSS-20B scores (Table \ref{tab:posealign_results}) are consistent with the GPT-5 results and both follow the same order, which means that these metrics are consistent and reliable.
\label{sec:gpt_5_llm_as_judge}

\paragraph{Full parameter tuning experiment on Chat-Scene and 3D-LLAVA.} We provide all performance metrics of tuning the clip ratio $X$ for two representative backbones, Chat-Scene and 3D-LLAVA, to examine the effect of $X$ on directional alignment. As shown in Tables~\ref{tab:parameter_tuning_chat_scene} and~\ref{tab:parameter_tuning_3d_llava}, performance first improves with moderate clip ratios and then gradually declines beyond $X=0.3$, exhibiting a clear unimodal trend except on Scan2Cap. Extremely small values ($X{<}0.1$) cause overfitting to marginal pose deviations, while large values ($X{>}0.4$) discard too many candidate poses, reducing diversity and robustness. Both models achieve their best balance between stability and discriminability near $X{=}0.3$, where directional understanding improves across ScanQA, SQA3D, and Scan2Cap without sacrificing general reasoning ability. This consistency across two architectures validates the robustness of the proposed cut-ratio mechanism and supports the choice of $X{=}0.3$ as the default setting in all main experiments.

\section{Prompt Designs}
\label{sec:prompts}
\paragraph{Template for Direction-critical question analysis.} The following template is for ScanQA dataset, while the prompts for other datasets are similarly constructed with different examples but the same prompt structure.

\label{sec:appendix_direction_gpt_prompt}

\begin{lstlisting}
Your job is to look at a question asking about the interior of a room and a bunch of associated ground truth answers, and then assign the problem as either [A: "NEED_LATERAL_DIRECTION", B: "DO_NOT_NEED_LATERAL_DIRECTION"] in order to answer.
First, I will give examples of each class, and then you will classify a new question.

The following are examples of DO_NOT_NEED_LATERAL_DIRECTION questions.
```
Question 1: What color is the chair in the kitchen?
Answer 1: "brown", "dark brown".
Question 2: What is the top of the table?
Answer 2: "tv", "television".
Question 3: What is placed next to the fridge?
Answer 3: "door", "the beige door".
Question 4: Where are the two chairs in front of one another?
Answer 4: "end table closest to door", "across from each other over table"
```
These questions all DO_NOT_NEED_LATERAL_DIRECTION because:
    - Either they ask a different trait of the object than its direction (e.g., color, material, shape, size, etc.).
    - Or the direction is dependent on an absolute direction (e.g., "top", "bottom", "next to", "between", "end") which can be implied without additional direction information.
    - Or like Question 4 where the reference frame is not needed or can be implied despite the existence of directional words like "in front of", because two chairs facing each other will be facing each other no matter where you look at them from.
    - While the question may contain spatial words, it does not necessarily need the directional information to answer.

The following are examples of question that NEED_LATERAL_DIRECTION.
```
Question 1: How many brown chairs are on the left of the brown table?
Answer 1: "4", "4"
Question 2: Where is the tall chair on the right of the table?
Answer 2: "to left of narrow table tv", "next to 2 other chairs as third chair on right"
Question 3: What part of the room has a fridge to its right?
Answer 3: "right corner wall adjacent to tv", "corner with tv alongside same wall"
Question 4: Where is the beige wooden desk placed?
Answer 4: "up against wall", "at front of class"
Question 5: What is on the front of the brown table?
Answer 5: "tv", "chair"
```
These predicted answers all NEED_LATERAL_DIRECTION because:
    - A relative direction word (e.g., "left", "right", "front", "back", "against") is present in the question or any of the answers.
    - Even if the question asks about directions relative to an object, in some cases the reference frame still cannot be implied, e.g., cylindrical or symmetric objects like dishes, tables, beds, etc, just like Question 5 above.
    - The question may be asking about a different trait of the object than its direction (e.g., color, material, shape, size, etc.), but the question cannot be answered without the directional information.
    
Also note the following things:
- For classifying the questions where the answers or question may seem odd, ignore that and just focus on whether the question can be answered without directional information.
- The answers may not coincide with each other, but that is okay. There can be multiple objects to the left or right of something. Just focus on whether the question can be answered without directional information.

Here is a new question. Simply reply with either NEED_LATERAL_DIRECTION or DO_NOT_NEED_LATERAL_DIRECTION. Do not ask any more questions to the user.
```
Question: {question}
Answers: {gt_answer}
```

Classify the question as one of:
A: NEED_LATERAL_DIRECTION
B: DO_NOT_NEED_LATERAL_DIRECTION

Just return the letters "A" or "B", with no text around it.

\end{lstlisting}

\paragraph{Template for LLM-as-judge.} The following template is for LLM-as-judge on the result of ScanQA dataset, while the prompts for other datasets are similarly constructed with different examples but the same prompt structure. The templates can be found in the open-sourced code. 

\begin{lstlisting}
Your job is to look at a question, a ground-truth answer, and a predicted answer, and then assign a grade of either ["CORRECT", "INCORRECT"].
First, I will give examples of each grade, and then you will grade a new example.

The following are examples of CORRECT predicted answers.
```
Question 1: What type of stool is to the left of the door?
Ground truth: "bar stool", "black topped wooden stool"
Predicted answer 1: wooden stool.
Predicted answer 2: black topped bar stool.
Predicted answer 3: wooden bar stool.

Question 2: Where is the door of the refrigerator located?
Ground truth: "next to microwave", "on refrigerator to right of door"
Predicted answer 1: next to the microwave.
Predicted answer 2: on the refrigerator to the right of the door.

Question 3: What is in the bin next to the door?
Ground Truth: "trash for recycling", "recycling bins"
Predicted answer 1: trash can.
Predicted answer 2: litter waiting for recycling.
Predicted answer 3: waste.
```
These predicted answers are all CORRECT because:
    - They fully contain the important information in one of the ground truths. Aliases and synonyms are acceptable.
    - They do not contain any information that contradicts the ground truth.
    - Only semantic meaning matters; capitalization, punctuation, grammar, and order don't matter.
    - Hedging and guessing are permissible, provided that the ground truth is fully included and the response contains no incorrect information or contradictions.

The following are examples of INCORRECT predicted answers.
```
Question 1: What is to the right of the refrigerator and above the kitchen cabinet?
Ground truth: "microwave", "microwave"
Predicted answer 1: kitchen counter.
Predicted answer 2: kitchen sink.
Predicted answer 3: water tap.

Question 2: Where is the door of the refrigerator located?
Ground truth: "next to microwave", "on refrigerator to right of door"
Predicted answer 1: on the microwave.
Predicted answer 2: on the door next to the refrigerator.
Predicted answer 3: to right of refrigerator.
Predicted answer 4: on the microwave above the refrigerator.
Predicted answer 5: on the refrigerator to left of the door.

Question 3: How much larger is the blue recycling bin compared to the beige trash can?
Ground truth: "twice as wide", "about double size"
Predicted answer 1: larger.
Predicted answer 2: blue recycling bin is larger than beige trash can.
Predicted answer 3: blue recycling bin is much larger than beige trash can.
```
These predicted answers are all INCORRECT because:
    - None of the information in the ground truth is included in the answer.
    - Or the predicted answer contradicts facts contained in the question and the ground truth. Pay extra attention to object relations, e.g., the door of the refrigerator cannot be on the microwave or another door. Having the correct words is not enough; the meaning must be correct.
    - The answer shies away from the question and does not provide any useful information.

Also note the following things:
- For grading questions where the ground truth contains a number, the predicted answer needs to be correct to the counting figure in the ground truth, but can be relaxed by +-1 when describing rough figures like relative size. For example, consider a question "How much larger is the brown cabinet compared to the purple stool?" with ground truth "5 times larger". 
    - Predicted answers "5", "4", and "6" are all CORRECT. 
    - Predicted answers "2", "3" and "8" are INCORRECT. 
    - Predicted answers "larger" and "more than 2" are considered INCORRECT because they provide no useful information.
- Do not punish predicted answers if they omit information that would be clearly inferred from the question.
    - For example, consider the question "What surface texture is the wooden cardboard?" and the gold target "wooden matte", "brown flat". The predicted answer "flat" would be considered CORRECT, even though it does not include "wooden", since it is clear from the question that the cardboard is wooden.

Here is a new example. Simply reply with either CORRECT or INCORRECT.
```
Question: {question}
Ground truth: {ground_truth}
Predicted answer: {predicted_answer}
```

Grade the predicted answer of this new question as one of:
A: CORRECT
B: INCORRECT

Just return the letters "A" or "B", with no text around it.
\end{lstlisting}